\documentclass{article}

\usepackage{times}
\usepackage{graphicx} 
\usepackage{subfigure}

\usepackage{natbib}

\usepackage{algorithm}
\usepackage{algorithmic}

\usepackage{hyperref}

\usepackage{amsmath}
\usepackage{amssymb}
\usepackage{pinyin}
\usepackage{CJK}

\usepackage[accepted]{icml2015}

\newcommand\blfootnote[1]{%
  \begingroup
  \renewcommand\thefootnote{}\footnote{#1}%
  \addtocounter{footnote}{-1}%
  \endgroup
}

\icmltitlerunning{Text Understanding from Scratch}

\begin{document}
\begin{CJK}{UTF8}{gbsn}

%
%
%
%
%
%

\twocolumn[
\icmltitle{Text Understanding from Scratch}

\icmlauthor{Xiang Zhang}{xiang@cs.nyu.edu}
\icmlauthor{Yann LeCun}{yann@cs.nyu.edu}
\icmladdress{Computer Science Department, Courant Institute of Mathematical Sciences, New York University}

\icmlkeywords{text understanding, convolutional networks}

\vskip 0.3in
]

\blfootnote{This technical report is superseded by a paper entitled ``Character-level Convolutional Networks for Text Classification'', arXiv:1509.01626. It has considerably more experimental results and a rewritten introduction.}

\begin{abstract} 
This article demonstrates that we can apply deep learning to text understanding from character-level inputs all the way up to abstract text concepts, using temporal convolutional networks\cite{LBBH98} (ConvNets). We apply ConvNets to various large-scale datasets, including ontology classification, sentiment analysis, and text categorization. We show that temporal ConvNets can achieve astonishing performance without the knowledge of words, phrases, sentences and any other syntactic or semantic structures with regards to a human language. Evidence shows that our models can work for both English and Chinese.
\end{abstract} 

\section{Introduction}
\label{sect:intro}

Text understanding consists in reading texts formed in natural languages, determining the explicit or implicit meaning of each elements such as words, phrases, sentences and paragraphs, and making inferences about the implicit or explicit properties of these texts\cite{N87}. This problem has been traditionally difficult because of the extreme variability in language formation\cite{L82}. To date,  most ways to handle text understanding, be it a hand-crafted parsing program or a statistically learnt model, have been resorted to the means of matching words statistics.

So far, most machine learning approaches to text understanding consist in tokenizing a string of characters into structures such as words, phrases, sentences and paragraphs, and then apply some statistical classification algorithm onto the statistics of such structures\cite{S01}. These techniques work well enough when applied to a narrowly defined domain, but the prior knowledge required is not cheap -- they need to pre-define a dictionary of interested words, and the structural parser needs to handle many special variations such as word morphological changes and ambiguous chunking. These requirements make text understanding more or less specialized to a particular language -- if the language is changed, many things must be engineered from scratch.

With the advancement of deep learning and availability of large datasets, methods of handling text understanding using deep learning techniques have gradually become available. One technique which draws great interests is word2vec\cite{MSCCD13}. Inspired by traditional language models, this technique constructs representation of words into a vector of fixed length trained under a large corpus. Based on the hope that machines may make sense of languages in a formal fashion, many researchers have tried to train a neural network for understanding texts based the features extracted from it or similar techniques, to name a few, \cite{FCGSBDM13}\cite{GHYD13}\cite{LM14}\cite{MLS13}\cite{PSM14}. Most of these techniques try to apply word2vec or similar techniques with an engineered language model.

On the other hand, some researchers have also tried to train a neural network from word level with little structural engineering\cite{CWB11}\cite{K14}\cite{JZ14}\cite{SG14}. In these works, a word level feature extractor such as lookup table\cite{CWB11} or word2vec\cite{MSCCD13} is used to feed a temporal ConvNet\cite{LBBH98}. After training, ConvNets worked for both structured prediction tasks such as  part-of-speech tagging and named entity recognition, and text understanding tasks such as sentiment analysis and sentence classification. They claim good results for various tasks, but the datasets and models are relatively small and there are still some engineered layers to represent structures such as words, phrases and sentences.

In this article we show that text understanding can be handled by a deep learning system without artificially embedding knowledge about words, phrases, sentences or any other syntactic or semantic structures associated with a language. We apply temporal ConvNets\cite{LBBH98} to various large-scale text understanding tasks, in which the inputs are quantized characters and the outputs are abstract properties of the text. Our approach is one that `learns from scratch', in the following 2 senses
\begin{enumerate}
\item ConvNets do not require knowledge of words -- working with characters is fine. This renders a word-based feature extractor (such as LookupTable\cite{CWB11} or word2vec\cite{MSCCD13}) unnecessary. All previous works start with words instead of characters, which is difficult to apply a convolutional layer directly due to its high dimension.
\item ConvNets do not require knowledge of syntax or semantic structures -- inference directly to high-level targets is fine. This also invalidates the assumption that structured predictions and language models are necessary for high-level text understanding.
\end{enumerate}

Our approach is partly inspired by ConvNet's success in computer vision. It has outstanding performance in various image recognition tasks\cite{GDDM13}\cite{KSH12}\cite{SEZMFL13}. These successful results usually involve some end-to-end ConvNet model that learns hierarchical representation from raw pixels\cite{GDDM13}\cite{ZF14}. Similarly, we hypothesize that when trained from raw characters, temporal ConvNet is able to learn the hierarchical representations of words, phrases and sentences in order to understand text.

\section{ConvNet Model Design}

In this section, we introduce the design of ConvNets for text understanding. The design is modular, where the gradients are obtained by back-propagation\cite{RHW86} to perform optimization.

\subsection{Key Modules}

The main component in our model is the temporal convolutional module, which simply computes a 1-D convolution between input and output. Suppose we have a discrete input function \(g(x) \in [1,l] \rightarrow \mathbb{R}\) and a discrete kernel function \(f(x) \in [1, k] \rightarrow \mathbb{R}\). The convolution \(h(y) \in [1,\lfloor (l-k)/d \rfloor + 1] \rightarrow \mathbb{R}\)  between \(f(x)\) and \(g(x)\) with stride \(d\) is defined as
\[
h(y) = \sum_{x = 1}^{k} f(x) \cdot g(y \cdot d - x + c),
\]
where \(c = k - d + 1\) is an offset constant. Just as in traditional convolutional networks in vision, the module is parameterized by a set of such kernel functions \(f_{ij} (x) ~ (i = 1, 2, \dots, m \text{ and } j = 1, 2, \dots, n )\) which we call \textit{weights}, on a set of inputs \(g_i(x) \) and outputs \(h_j(y)\). We call each \(g_i\) (or \(h_j\)) an input (or output) \textit{frame}, and \(m\) (or \(n\)) input (or output) frame size. The outputs \(h_j(y)\) is obtained by a sum over \(i\) of the convolutions between \(g_i(x)\) and \(f_{ij}(x)\).

One key module that helped us to train deeper models is temporal max-pooling. It is the same as spatial max-pooling module used in computer vision\cite{BBLP10}, except that it is in 1-D. Given a discrete input function \(g(x) \in [1,l] \rightarrow \mathbb{R}\), the max-pooling function \(h(y) \in [1,\lfloor (l-k)/d \rfloor + 1] \rightarrow \mathbb{R}\) of \(g(x)\) is defined as
\[
h(y) = \max_{x = 1}^{k} ~ g(y \cdot d - x + c),
\]
where \(c = k - d + 1\) is an offset constant. This very pooling module enabled us to train ConvNets deeper than 6 layers, where all others fail. The analysis by \cite{BPL10} might shed some light on this.

The non-linearity used in our model is the rectifier or thresholding function \(h(x) = \max\{0, x\}\), which makes our convolutional layers similar to rectified linear units (ReLUs)\cite{NH10}. We always apply this function after a convolutional or linear module, therefore we omit its appearance in the following. The algorithm used in training our model is stochastic gradient descent (SGD) with a minibatch of size 128, using momentum\cite{P64}\cite{SMDH13} \(0.9\) and initial step size \(0.01\) which is halved every 3 epoches for 10 times. The training method and parameters apply to all of our models. Our implementation is done using Torch 7\cite{CKF11}.

\subsection{Character quantization}

Our model accepts a sequence of encoded characters as input. The encoding is done by prescribing an alphabet of size \(m\) for the input language, and then quantize each character using 1-of-\(m\) encoding. Then, the sequence of characters is transformed to a sequence of such \(m\) sized vectors with fixed length \(l\). Any character exceeding length \(l\) is ignored, and any characters that are not in the alphabet including blank characters are quantized as all-zero vectors. Inspired by how long-short term memory (LSTM)\cite{HS97} work, we quantize characters in backward order. This way, the latest reading on characters is always placed near the beginning of the output, making it easy for fully connected layers to associate correlations with the latest memory. The input to our model is then just a set of frames of length \(l\), and the frame size is the alphabet size \(m\).

\begin{figure}[ht]
  \centering
  \subfigure[Binary]{\includegraphics[width=0.2\columnwidth]{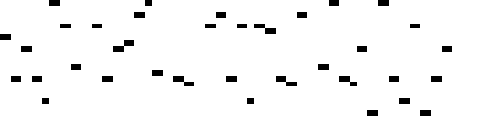}}
  \subfigure[Braille]{\includegraphics[width=0.7\columnwidth]{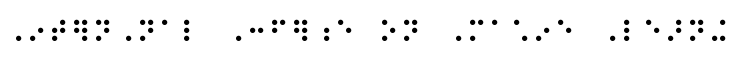}}
  \caption{Comparison of our binary encoding and Braille on the text ``International Conference on Machine Learning''}
  \label{fig:brai}
\end{figure}

One interesting thing about this quantization is that visually it is quite similar to Braille\cite{B1829} used for assisting blind reading, except that our encoding is more compact. Figure \ref{fig:brai} depicts this fact. It seems that when trained properly, humans can learn to read binary encoding of languages. This offers interesting insights and inspiration to why our approach could work.

The alphabet used in all of our models consists of 70 characters, including 26 English letters, 10 digits, new line and 33 other characters. They include:
\begin{verbatim}
abcdefghijklmnopqrstuvwxyz0123456789
-,;.!?:'''/\|_@#$%^&*~`+-=<>()[]{}
\end{verbatim}

Before feeding the input to our model, no normalization is done. This is because the input is already quite sparse by itself, with many zeros scattered around. Our models can learn from this simple quantization without problems.

\subsection{Model Design}

\begin{figure}[ht]
  \centering
  \includegraphics[width=\columnwidth]{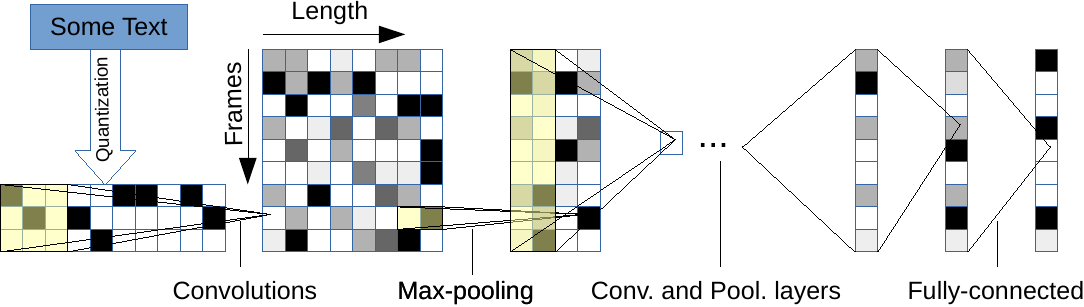}
  \caption{Illustration of our model}
  \label{fig:modl}
\end{figure}

We designed 2 ConvNets -- one large and one small. They are both 9 layers deep with 6 convolutional layers and 3 fully-connected layers, with different number of hidden units and frame sizes. Figure \ref{fig:modl} gives an illustration.

The input have number of frames equal to 69 due to our character quantization method, and the length of each frame is dependent on the problem. We also insert 2 dropout\cite{HSKSS12} modules in between the 3 fully-connected layers to regularize. They have dropout probability of 0.5. Table \ref{tab:conv} lists the configurations for convolutional layers, and table \ref{tab:full} lists the configurations for fully-connected (linear) layers.

\begin{table}[ht]
  \caption{Convolutional layers used in our experiments. The convolutional layers do not use stride and pooling layers are all non-overlapping ones, so we omit the description of their strides.}
  \label{tab:conv}
  \begin{center}
    \begin{tabular}{ccccc}
      \hline
      \abovespace\belowspace
      Layer & Large Frame & Small Frame & Kernel & Pool \\
      \hline
      \abovespace
      1 & 1024 & 256 & 7 & 3 \\
      2 & 1024 & 256 & 7 & 3 \\
      3 & 1024 & 256 & 3 & N/A \\
      4 & 1024 & 256 & 3 & N/A \\
      5 & 1024 & 256 & 3 & N/A \\
      \belowspace
      6 & 1024 & 256 & 3 & 3 \\
      \hline
    \end{tabular}
  \end{center}
\end{table}

Before starting training the models, we randomize the weights using Gaussian distributions. The mean and standard deviation used for initializing the large model is \((0, 0.02)\), and small model \((0, 0.05)\).

\begin{table}[ht]
  \caption{Fully-connected layers used in our experiments. The number of output units for the last layer is determined by the problem. For example, for a 10-class classification problem it will be 10.}
  \label{tab:full}
  \begin{center}
    \begin{tabular}{ccccc}
      \hline
      \abovespace\belowspace
      Layer & Output Units Large & Output Units Small \\
      \hline
      \abovespace
      7 & 2048 & 1024 \\
      8 & 2048 & 1024 \\
      \belowspace
      9 & \multicolumn{2}{c} {Depends on the problem} \\
      \hline
    \end{tabular}
  \end{center}
\end{table}

For different problems the input lengths are different, and so are the frame lengths. From our model design, it is easy to know that given input length \(l_0\), the output frame length after the last convolutional layer (but before any of the fully-connected layers) is \(l_6 = (l_0 - 96) / 27\). This number multiplied with the frame size at layer 6 will give the input dimension the first fully-connected layer accepts.

\subsection{Data Augmentation using Thesaurus}

Many researchers have found that appropriate data augmentation techniques are useful for controlling generalization error for deep learning models. These techniques usually work well when we could find appropriate invariant properties that the model should possess. For example, in image recognition a model should have some controlled invariance towards changes in translating, scaling, rotating and flipping of the input image. Similarly, in speech recognition we usually augment data by adding artificial noise background and changing the tone or speed of speech signal\cite{HCCCDEPSSCN14}.

In terms of texts, it is not reasonable to augment the data using signal transformations as done in image or speech recognition, because the exact order of characters may form rigorous syntactic and semantic meaning. Therefore, the best way to do data augmentation would have been using human rephrases of sentences, but this is unrealistic and expensive due the large volume of samples in our datasets. As a result, the most natural choice in data augmentation for us is to replace words or phrases with their synonyms.

We experimented data augmentation by using an English thesaurus, which is obtained from the \texttt{mytheas} component used in LibreOffice\footnote{\url{http://www.libreoffice.org/}} project. That thesaurus in turn was obtained from WordNet\cite{F05}, where every synonym to a word or phrase is ranked by the semantic closeness to the most frequently seen meaning.

To do synonym replacement for a given text, we need to answer 2 questions: which words in the text should be replaced, and which synonym from the thesaurus should be used for the replacement. To decide on the first question, we extract all replaceable words from the given text and randomly choose \(r\) of them to be replaced. The probability of number \(r\) is determined by a geometric distribution with parameter \(p\) in which \(P[r] \sim p^r\). The index \(s\) of the synonym chosen given a word is also determined by a another geometric distribution in which \(P[s] \sim q^s\). This way, the probability of a synonym chosen becomes smaller when it moves distant from the most frequently seen meaning.

It is worth noting that models trained using our large-scale datasets hardly require data augmentation, since their generalization errors are already pretty good. We will still report the results using this new data augmentation technique with \(p = 0.5\) and \(q = 0.5\).

\subsection{Comparison Models}

Since we have constructed several large-scale datasets from scratch, there is no previous publication for us to obtain a comparison with other methods. Therefore, we also implemented two fairly standard models using previous methods: the bag-of-words model, and a bag-of-centroids model via word2vec\cite{MSCCD13}.

The bag-of-words model is pretty straightforward. For each dataset, we count how many times each word appears in the training dataset, and choose 5000 most frequent ones as the bag. Then, we use multinomial logistic regression as the classifier for this bag of features.

As for the word2vec model, we first ran k-means on the word vectors learnt from Google News corpus with \(k = 5000\), and then use a bag of these centroids for multinomial logistic regression. This model is quite similar to the bag-of-words model in that the number of features is also 5000.

One difference between these two models is that the features for bag-of-words model are different for different datasets, whereas for word2vec they are the same. This could be one reason behind the phenomenon that bag-of-words consistently out-performs word2vec in our experiments. It might also be the case that the hope for linear separability of word2vec is not valid at all. That being said, our own ConvNet models consistently out-perform both.

\section{Datasets and Results}

In this part we show the results obtained from various datasets. The unfortunate fact in literature is that there is no openly accessible dataset that is large enough or with labels of sufficient quality for us, although the research on text understanding has been conducted for tens of years. Therefore, we propose several large-scale datasets, in hopes that text understanding can rival the success of image recognition when large-scale datasets such as ImageNet\cite{DDSLLF09} became available.

\subsection{DBpedia Ontology Classification}

DBpedia is a crowd-sourced community effort to extract structured information from Wikipedia\cite{LIJJKMHMKAB114}. The English version of the DBpedia knowledge base provides a consistent ontology, which is shallow and cross-domain. It has been manually created based on the most commonly used infoboxes within Wikipedia. Some ontology classes in DBpedia contain hundreds of thousands of samples, which are ideal candidates to construct an ontology classification dataset.

The DBpedia ontology classification dataset is constructed by picking 14 non-overlapping classes from DBpedia 2014. They are listed in table \ref{tab:dbpd}. From each of these 14 ontology classes, we randomly choose 40,000 training samples and 5,000 testing samples. Therefore, the total size of the training dataset is 560,000 and testing dataset 70,000. 


\begin{table}[ht]
  \caption{DBpedia ontology classes. The numbers contain only samples with both a title and a short abstract.}
  \label{tab:dbpd}
  \begin{center}
    \begin{tabular}{lccc}
      \hline
      \abovespace\belowspace
      Class & Total & Train & Test  \\
      \hline
      \abovespace
      Company & 63,058 & 40,000 & 5,000 \\
      Educational Institution & 50,450 & 40,000 & 5,000 \\
      Artist & 95,505 & 40,000 & 5,000 \\
      Athlete & 268,104 & 40,000 & 5,000 \\
      Office Holder & 47,417 & 40,000 & 5,000 \\
      Mean Of Transportation & 47,473 & 40,000 & 5,000 \\
      Building & 67,788 & 40,000 & 5,000 \\
      Natural Place & 60,091 & 40,000 & 5,000 \\
      Village & 159,977 & 40,000 & 5,000 \\
      Animal & 187,587 & 40,000 & 5,000 \\
      Plant & 50,585 & 40,000 & 5,000 \\
      Album & 117,683 & 40,000 & 5,000 \\
      Film & 86,486 & 40,000 & 5,000 \\
      \belowspace
      Written Work & 55,174 & 40,000 & 5,000 \\
      \hline
    \end{tabular}
  \end{center}
\end{table}

Before feeding the data to the models, we concatenate the title and short abstract together to form a single input for each sample. The length of input used was \(l_0 = 1014\), therefore the frame length after last convolutional layer is \(l_6 = 34\). Using an NVIDIA Tesla K40, Training takes about 5 hours per epoch for the large model, and 2 hours for the small model. Table \ref{tab:dbpr} shows the classification results.

\begin{table}[ht]
  \caption{DBpedia results. The numbers are accuracy.}
  \label{tab:dbpr}
  \begin{center}
    \begin{tabular}{cccc}
      \hline
      \abovespace\belowspace
      Model & Thesaurus & Train & Test  \\
      \hline
      \abovespace
      Large ConvNet & No & \textbf{99.96\%} & 98.27\% \\
      Large ConvNet & Yes & 99.89\% & \textbf{98.40\%} \\
      Small ConvNet & No & 99.37\% & 98.02\% \\
      Small ConvNet & Yes & 99.62\% & 98.15\% \\
      Bag of Words & No & 96.29\% & 96.19\% \\
      \belowspace
      word2vec & No & 89.32\% & 89.09\% \\
      \hline
    \end{tabular}
  \end{center}
\end{table}

The results from table \ref{tab:dbpr} indicate both good training and testing errors from our models, with some improvement from thesaurus augmentation. We believe this is a first evidence that a learning machine does not require knowledge about words, phrases, sentences, paragraphs or any other syntactical or semantic structures to understand text. That being said, we want to point out that ConvNets by their design have the capacity to learn such structured knowledge.

\begin{figure}[ht]
  \centering
  \includegraphics[width=\columnwidth]{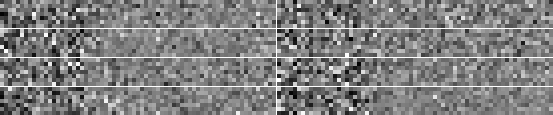}
  \caption{Visualization of first layer weights}
  \label{fig:1lay}
\end{figure}

Figure \ref{fig:1lay} is a visualization of some kernel weights in the first layer of the large model trained without thesaurus augmentation. Each block represents a randomly chosen kernel, with its horizontal direction iterates over input frames and vertical direction over kernel size. In the visualization, black (or white) indicates large negative (or positive) values, and gray indicates values near zero. It seems very interesting that the network has learnt to care more about the variations in letters than other characters. This phenomenon is observed in models for all of the datasets.

\subsection{Amazon Review Sentiment Analysis}

The purpose of sentiment analysis is to identify and extract subjective information in different kinds of source materials. This task, when presented with the text written by some user, could be formulated as a normal classification problem in which each class represents a degree indicator for user's subjective view. One example is the score system used from Amazon, which is a discrete score from 1 to 5 indicating user's subjective rating of a product. The rating usually comes with a review text, which is a valuable source for us to construct a sentiment analysis dataset.

We obtained an Amazon review dataset from the Stanford Network Analysis Project (SNAP), which spans 18 years with 34,686,770 reviews from 6,643,669 users on 2,441,053 products\cite{ML13}. This dataset contains review texts of extremely variate character lengths from 3 to 32,788, in which the mean is around 764. To construct a sentiment analysis dataset, we chose review texts with character lengths between 100 and 1014. Apart from constructing from the original 5 score labels, we also construct a sentiment polarity dataset in which labels 1 and 2 are converted to negative and 4 and 5 positive. There are also large number of duplicated reviews in which the title and review text are the same. We removed these duplicates. Table \ref{tab:amzd} lists the number of samples for each score and the number sampled for the 2 dataset.

\begin{table}[ht]
  \caption{Amazon review datasets. Column ``total'' is the total number of samples for each score. Column ``full'' and ``polarity'' are number of samples chosen for full score dataset and polarity dataset, respectively.}
  \label{tab:amzd}
  \begin{center}
    \begin{tabular}{ccccc}
      \hline
      \abovespace\belowspace
       & Total & Full & Polarity  \\
      \hline
      \abovespace
      1 & 2,746,559 & 730,000 & 1,275,000 \\
      2 & 1,791,219 & 730,000 & 725,000 \\
      3 & 2,892,566 & 730,000 & 0 \\
      4 & 6,551,166 & 730,000 & 725,000 \\
      \belowspace
      5 & 20,705,260 & 730,000 & 1,275,000 \\
      \hline
    \end{tabular}
  \end{center}
\end{table}


We ignored score 3 for polarity dataset because some texts in that score are not obviously negative or positive. Many researchers have shown that with some random text, the inter-rater consensus on polarity is only about 60\% - 80\%\cite{GA05}\cite{KH04}\cite{SM08}\cite{VG60}\cite{WWB01}\cite{WWH05}. We believe that by picking out score 3, the labels would have higher quality with a clearer indication of positivity or negativity. We could have included a third ``neutral'' class, but that would significantly reduce the number of samples for each class since sample imbalance is not desirable.

For the full score dataset, we randomly selected 600,000 samples for each score for training and 130,000 samples for testing. The size of training set is then 3,000,000 and testing 650,000. For the polarity dataset, we randomly selected 1,800,000 samples for each positive or negative label as training set and 200,000 samples for testing. In total, the polarity dataset has 3,600,000 training samples and 400,000 testing samples.

\begin{table}[ht]
  \caption{Result on Amazon review full score dataset. The numbers are accuracy.}
  \label{tab:amzr}
  \begin{center}
    \begin{tabular}{cccc}
      \hline
      \abovespace\belowspace
      Model & Thesaurus & Train & Test  \\
      \hline
      \abovespace
      Large ConvNet & No & 62.96\% & 58.69\% \\
      Large ConvNet & Yes & 68.90\% & 59.55\% \\
      Small ConvNet & No & \textbf{69.24\%} & 59.47\% \\
      Small ConvNet & Yes & 62.11\% & \textbf{59.57\%} \\
      Bag of Words & No & 54.45\% & 54.17\% \\
      \belowspace
      word2vec & No & 36.56\% & 36.50\% \\
      \hline
    \end{tabular}
  \end{center}
\end{table}

\begin{table}[ht]
  \caption{Result on Amazon review polarity dataset. The numbers are accuracy.}
  \label{tab:amzp}
  \begin{center}
    \begin{tabular}{cccc}
      \hline
      \abovespace\belowspace
      Model & Thesaurus & Train & Test  \\
      \hline
      \abovespace
      Large ConvNet & No & \textbf{97.57\%} & 94.49\% \\
      Large ConvNet & Yes & 96.82\% & \textbf{95.07\%} \\
      Small ConvNet & No & 96.03\% & 94.50\% \\
      Small ConvNet & Yes & 95.44\% & 94.33\% \\
      Bag of Words & No & 89.96\% & 89.86\% \\
      \belowspace
      word2vec & No & 72.95\% & 72.86\% \\
      \hline
    \end{tabular}
  \end{center}
\end{table}

Because we limit the maximum length of the text to be 1014, we can safely set the input length to be 1014 and use the same configuration as the DBpedia model. Models for Amazon review datasets took significantly more time to go over each epoch. The time taken for the large model per epoch is about a 5 days, and small model 2 days, with the polarity training taking a little bit longer. Table \ref{tab:amzr} and table \ref{tab:amzp} list the results on full score dataset and polarity dataset, respectively.

\begin{figure}[ht]
  \centering
  \subfigure[Train]{\includegraphics[width=0.35\columnwidth]{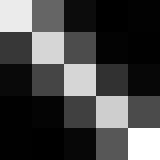}}
  \subfigure[Test]{\includegraphics[width=0.35\columnwidth]{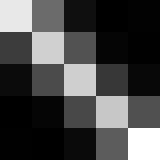}}
  \caption{Confusion matrices on full score Amazon Review prediction. White values are 1 and black 0. Vertical direction iterates over true score from top to bottom, and horizontal direction iterates over predicted scores from left to right.}
  \label{fig:amzc}
\end{figure}

It seems that our models work much better on the polarity dataset than the full score dataset. This is to be expected, since full score prediction means more confusion between nearby score labels. To demonstrate this, figure \ref{fig:amzc} shows the training and testing confusion matrices.

\subsection{Yahoo! Answers Topic Classification}

Yahoo! Answers is a web site where people post questions and answers, all of which are public to any web user willing to browse or download them. We obtained Yahoo! Answers Comprehensive Questions and Answers version 1.0 dataset through the Yahoo! Webscope program. The data they have collected is the Yahoo! Answers corpus as of October 25th, 2007. It includes all the questions and their corresponding answers. The corpus contains 4,483,032 questions and their answers. In addition to question and answer text, the corpus contains a small amount of metadata, i.e., which answer was selected as the best answer, and the category and sub-category that was assigned to each question.

We constructed a topic classification dataset from this corpus using 10 largest main categories. They are listed in table \ref{tab:yadb}. Each class contains 140,000 training samples and 6,000 testing samples. Therefore, the total number of training samples is 1,400,000 and testing samples 60,000 in this dataset. From all the answers and other meta-information, we only used the best answer content and the main category information. 

\begin{table}[ht]
  \caption{Yahoo! Answers topic classification dataset}
  \label{tab:yadb}
  \begin{center}
    \begin{tabular}{lcccc}
      \hline
      \abovespace\belowspace
      Category & Total & Train & Test \\
      \hline
      \abovespace
      Society \& Culture & 295,340 & 140,000 & 6,000 \\
      Science \& Mathematics & 169,586 & 140,000 & 6,000 \\
      Health & 278,942 & 140,000 & 6,000 \\
      Education \& Reference & 206,440 & 140,000 & 6,000 \\
      Computers \& Internet & 281,696 & 140,000 & 6,000 \\
      Sports & 146,396 & 140,000 & 6,000 \\
      Business \& Finance & 265,182 & 140,000 & 6,000 \\
      Entertainment \& Music & 440,548 & 140,000 & 6,000 \\
      Family \& Relationships & 517,849 & 140,000 & 6,000 \\
      \belowspace
      Politics \& Government & 152,564 & 140,000 & 6,000 \\
      \hline
    \end{tabular}
  \end{center}
\end{table}

The Yahoo! Answers dataset also contains questions and answers of various lengths, up to 4000 characters. During training we still set the input length to be 1014 and truncate the rest if necessary. But before truncation, we concatenated the question title, question content and best answer content in reverse order so that the question title and content are less likely to be truncated. It takes about 1 day for one epoch on the large model, and about 8 hours for the small model. Table \ref{tab:yadr} details the results on this dataset.

\begin{table}[ht]
  \caption{Results on Yahoo! Answers dataset. The numbers are accuracy.}
  \label{tab:yadr}
  \begin{center}
    \begin{tabular}{cccc}
      \hline
      \abovespace\belowspace
      Model & Thesaurus & Train & Test  \\
      \hline
      \abovespace
      Large ConvNet & No & 73.42\% & 70.45\% \\
      Large ConvNet & Yes & \textbf{75.55\%} & \textbf{71.10\%} \\
      Small ConvNet & No & 72.84\% & 70.16\% \\
      Small ConvNet & Yes & 72.51\% & 70.16\% \\
      Bag of Words & No & 66.83\% & 66.62\% \\
      \belowspace
      word2vec & No & 56.37\% & 56.47\% \\
      \hline
    \end{tabular}
  \end{center}
\end{table}

One interesting thing from the results on Yahoo! Answers dataset is that both training and testing accuracy values are quite small compared to the results we obtained from other datasets, whereas the generalization error is pretty good. One hypothesis for this is that there are some intrinsic confusions in determining between some classes given a pair of question and answer.

\begin{figure}[ht]
  \centering
  \subfigure[Train]{\includegraphics[width=0.4\columnwidth]{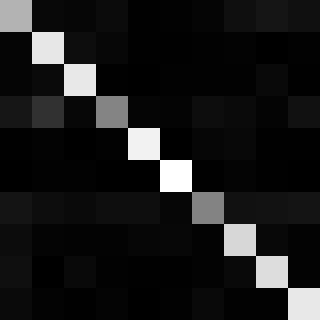}}
  \subfigure[Test]{\includegraphics[width=0.4\columnwidth]{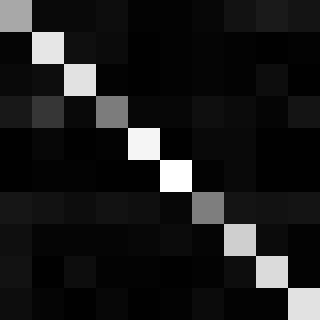}}
  \caption{Confusion matrices on Yahoo! Answers dataset. White values are 1 and black 0. Vertical direction iterates over true classes from top to bottom, and horizontal direction iterates over predicted classes from left to right.}
  \label{fig:yadc}
\end{figure}

Figure \ref{fig:yadc} shows the confusion matrix for the large model without thesaurus augmentation. It indicates relatively large confusion for classes ``Society \& Culture'', ``Education \& Reference'', and ``Business \& Finance''.

\subsection{News Categorization in English}

News is one of the largest parts of the entire web today, which makes it a good candidate to build text understanding models. We obtained the AG's corpus of news article on the web\footnote{\url{http://www.di.unipi.it/~gulli/AG_corpus_of_news_articles.html}}. It contains 496,835 categorized news articles from more than 2000 news sources. We choose 4 largest categories from this corpus to construct our dataset, using only the \textit{title} and \textit{description} fields.

\begin{table}[ht]
  \caption{AG's news corpus. Only categories used are listed.}
  \label{tab:nwsd}
  \begin{center}
    \begin{tabular}{lcccc}
      \hline
      \abovespace\belowspace
      Category & Total & Train & Test \\
      \hline
      \abovespace
      World & 81,456 & 30,000 & 1,900 \\
      Sports & 62,163 & 30,000 & 1,900 \\
      Business & 56,656 & 30,000 & 1,900 \\
      \belowspace
      Sci/Tech & 41,194 & 30,000 & 1,900 \\
      \hline
    \end{tabular}
  \end{center}
\end{table}

Table \ref{tab:nwsd} is a summary of the dataset. From each category, we randomly chose 30,000 samples as training and 1,900 as testing. The total number of training samples is then 120,000 and testing 7,600. Compared to other datasets we have constructed, this dataset is relatively small. Therefore the time taken for one epoch using the large model is only 3 hours, and about 1 hour for the small model. 

\begin{table}[ht]
  \caption{Result on AG's news corpus. The numbers are accuracy}
  \label{tab:nwsr}
  \begin{center}
    \begin{tabular}{cccc}
      \hline
      \abovespace\belowspace
      Model & Thesaurus & Train & Test  \\
      \hline
      \abovespace
      Large ConvNet & No & 99.44\% & \textbf{87.18\%} \\
      Large ConvNet & Yes & \textbf{99.49\%} & 86.61\% \\
      Small ConvNet & No & 99.20\% & 84.35\% \\
      Small ConvNet & Yes & 96.81\% & 85.20\% \\
      Bag of Words & No & 88.02\% & 86.69\% \\
      \belowspace
      word2vec & No & 78.20\% & 76.73\% \\
      \hline
    \end{tabular}
  \end{center}
\end{table}

Similarly as our previous experiments, we also use an input length of 1014 for this dataset after title and description are concatenated. The actual resulting maximum length of all the inputs is 9843, but the mean is only around 232.

Table \ref{tab:nwsr} lists the results. It shows a sign of overfitting from our models, which suggests that to achieve good text understanding results ConvNets require a large corpus in order to learn from scratch.

\subsection{News Categorization in Chinese}

One immediate advantage from our dictionary-free design is its applicability to other kinds of human languages. Our simple approach only needs an alphabet of the target language using one-of-n encoding. For languages such as Chinese, Japanese and Korean where there are too many characters, one can simply use its romanized (or latinized) transcription and quantize them just like in English. Better yet, the romanization or latinization is usually phonemic or phonetic, which rivals the success of deep learning in speech recognition\cite{HCCCDEPSSCN14}. Here we investigate one example: news categorization in Chinese.

The dataset we obtained consists of the SogouCA and SogouCS news corpora\cite{WZMR08}, containing in total 2,909,551 news articles in various topic channels. Among them, about 2,644,110 contain both a title and some content. We then labeled the each piece of news using its URL, by manually classify the their domain names. This gives us a large corpus of news articles labeled with their categories. There are a large number categories but most of them contain only few articles. We choose 5 categories -- ``sports'', ``finance'', ``entertainment'', ``automobile'' and ``technology''. The number of training samples selected for each class is 90,000 and testing 12,000, as table \ref{tab:sogd} shows.

\begin{table}[ht]
  \caption{Sogou News dataset}
  \label{tab:sogd}
  \begin{center}
    \begin{tabular}{lcccc}
      \hline
      \abovespace\belowspace
      Category & Total & Train & Test \\
      \hline
      \abovespace
      Sports & 645,931 & 90,000 & 12,000 \\
      Finance & 315,551 & 90,000 & 12,000 \\
      Entertainment & 160,409 & 90,000 & 12,000 \\
      Automobile & 167,647 & 90,000 & 12,000 \\
      \belowspace
      Technology & 188,111 & 90,000 & 12,000 \\
      \hline
    \end{tabular}
  \end{center}
\end{table}

The romanization or latinization form we have used is Pinyin, which is a phonetic system for transcribing the Mandarin pronunciations. During this procedure, we used the \verb|pypinyin| package combined with \verb|jieba| Chinese segmentation system. The resulting Pinyin text had each tone appended their finals as numbers between 1 and 4. 

Similar as before, we concatenate title and content to form an input sample. The texts has a wide range of lengths from 14 to 810959. Therefore, during data acquisition procedure we constrain the length to stay between 100 and 1014 whenever possible. In the end, we also apply same models as before to this dataset, for which the input length is 1014. We ignored thesaurus augmentation for this dataset. Table \ref{tab:sogr} lists the results.

\begin{table}[ht]
  \caption{Result on Sogou News corpus. The numbers are accuracy}
  \label{tab:sogr}
  \begin{center}
    \begin{tabular}{cccc}
      \hline
      \abovespace\belowspace
      Model & Thesaurus & Train & Test  \\
      \hline
      \abovespace
      Large ConvNet & No & \textbf{99.14\%} & \textbf{95.12\%} \\
      Small ConvNet & No & 93.05\% & 91.35\% \\
      \belowspace
      Bag of Words & No & 92.97\% & 92.78\% \\
      \hline
    \end{tabular}
  \end{center}
\end{table}

The input for a bag-of-words model is obtained by considering each Pinyin at Chinese character level as a word. These results indicate consistently good performance from our ConvNet models, even though it is completely a different kind of human language. This is one evidence to our belief that ConvNets can be applied to any human language in similar ways for text understanding tasks.

\section{Outlook and Conclusion}

In this article we provide a first evidence on ConvNets' applicability to text understanding tasks from scratch, that is, ConvNets do not need any knowledge on the syntactic or semantic structure of a language to give good benchmarks text understanding. This evidence is in contrast with various previous approaches where a dictionary of words is a necessary starting point, and usually structured parsing is hard-wired into the model\cite{CWB11}\cite{K14}\cite{JZ14}\cite{SG14}.

Deep learning models have been known to have good representations across domains or problems, in particular for image recognition\cite{RASC14}. How good the learnt representations are for language modeling is also one interesting question to ask in the future. Beyond that, we can also consider how to apply unsupervised learning to language models learnt from scratch. Previous embedding methods\cite{CWB11}\cite{MSCCD13}\cite{LM14} have shown that predicting words or other patterns missing from the input could be useful. We are eager to see how to apply these transfer learning and unsupervised learning techniques with our models.

Recent research shows that it is possible to generate text description of images from the features learnt in a deep image recognition model, using either fragment embedding\cite{KJF14} or recurrent neural networks such as long-short term memory (LSTM)\cite{VTBE14}. The models in this article show very good ability for understanding natural languages, and we are interested in using the features from our model to generate a response sentence in similar ways. If this could be successful, conversational systems could have a big advancement.

It is also worth noting that natural language in its essence is time-series in disguise. Therefore, one natural extended application for our approach is towards time-series data, in which a hierarchical feature extraction mechanism could bring some improvements over the recurrent and regression models used widely today.

In this article we only apply ConvNets to text understanding for its semantic or sentiment meaning. One other apparent extension is towards traditional NLP tasks such as chunking, named entity recognition (NER) and part-of-speech (POS) tagging. To do them, one would need to adapt our models to structured outputs. This is very similar to the seminal work by Collobert and Weston\cite{CWB11}, except that we probably no longer need to construct a dictionary and start from words. Our work also makes it easy to extend these models to other human languages.

One final possibility from our model is learning from symbolic systems such as mathematical equations, logic expressions or programming languages. Zaremba and Sutskever\cite{ZS14} have shown that it is possible to approximate program executing using a recurrent neural network. We are also eager to see how similar projects could work out using our ConvNet models.

With so many possibilities, we believe that ConvNet models for text understanding could go beyond from what this article shows and bring important insights towards artificial intelligence in the future.

\section*{Acknowledgement}

We gratefully acknowledge the support of NVIDIA Corporation with the donation of 2 Tesla K40 GPUs used for this research.

\bibliography{article}
\bibliographystyle{icml2015}

\end{CJK}
\end{document}